\documentclass{llncs}

\usepackage[clean=true,width=12.2cm,inkscape={/opt/homebrew-cask/Caskroom/inkscape/0.48.5-2/Inkscape.app/Contents/Resources/bin/inkscape -z -D}]{svg}
\usepackage{hyperref}

\begin{document}

\title{Automatically\\Segmenting Oral History Transcripts}

\author{Ryan Shaw}

\institute{School of Information and Library Science \\
  University of North Carolina at Chapel Hill \\
  \email{ryanshaw@unc.edu}\\
  \url{https://aeshin.org}}

\maketitle

\begin{abstract}
Dividing oral histories into topically coherent segments can make them more accessible online. People regularly make judgments about where coherent segments can be extracted from oral histories. But making these judgments can be taxing, so automated assistance is potentially attractive to speed the task of extracting segments from open-ended interviews. When different people are asked to extract coherent segments from the same oral histories, they often do not agree about precisely where such segments begin and end. This low agreement makes the evaluation of algorithmic segmenters challenging, but there is reason to believe that for segmenting oral history transcripts, some approaches are more promising than others. The BayesSeg algorithm performs slightly better than TextTiling, while TextTiling does not perform significantly better than a uniform segmentation. BayesSeg might be used to suggest boundaries to someone segmenting oral histories, but this segmentation task needs to be better defined.
\end{abstract}

\section{Introduction}

Oral histories are rich and unique documents of the past and our memory of it. Putting oral histories on the web makes them more accessible, but they remain daunting to consume \cite{Frisch2006}. It requires a significant time commitment to listen to a one or two hour interview. This is why curators of oral history collections, when creating ``exhibits'' of their materials for the public, usually select short extracts from longer interviews. Scholars also select extracts from their recordings when presenting their work to a live audience \cite[265]{Thompson2000}. Short, self-contained extracts of oral histories are more approachable than the longer recordings from which they are taken, but selecting these extracts is labor-intensive.

The labor required to select extracts from long interviews could be reduced through partial automation. The selection of extracts can be modeled as a two-step process: first separate the interview into segments, then select the segments of interest. In this paper I focus on the first step: segmentation of interview transcripts. Automatic segmentation of discourse is an established problem for natural language processing. Research has shown that the efficacy of segmentation algorithms varies widely with text genre and the kinds of segments desired \cite[38]{Stede2011}. Thus one cannot assume that segmentation techniques that have proved effective for other genres of discourse, using different definitions of segment, will work well for finding potential extract boundaries in oral history interviews.  

I collected 829 judgments about oral history  boundaries. I then used their judgments to evaluate the performance of two automatic text segmentation algorithms, TextTiling \cite{Hearst1997} and BayesSeg \cite{Eisenstein2008}. The results showed that BayesSeg performs slightly better than TextTiling, while TextTiling does not perform significantly better than a uniform segmentation (all segments of equal length). On the basis of these results, I conclude that BayesSeg might be used to lower the costs of selecting extracts from oral histories, but that the segmentation task for oral histories needs to be better defined.

\section{Choosing the Telling Extract}

Oral historians see ``choosing the telling extract'' \cite[265]{Thompson2000} as an essential part of communicating to their audiences. The actual words of interviewees are more evocative that the historian's paraphrase, and extracts are seen as essential to providing ``texture'' and ``voice'' to the historian's interpretation \cite[27--28]{Lippincott2012}. While some historians will limit themselves to pulling brief quotes to illustrate their arguments, others yield far more to their respondents, aiming to let them ``tell their own story'' by stitching together long extracts. This is particularly common when the medium of communication allows for audio to be presented. When giving talks, oral historians will often build their presentation around a handful of extracts, four or five minutes each \cite[266]{Thompson2000}. Incorporating such extracts is now a possibility for authors of monographs as well, thanks to recent developments in e-books.

Libraries and archives have made oral history interviews considerably more accessible by digitizing  and publishing them on the web. A web-based finding aid typically lists the interviews available with descriptions of who was interviewed, when and where, and a set of subject headings indicating the content of the interview. Yet as Frisch \cite{Frisch2006} and others have argued, this kind of description at the interview level is by itself insufficient for making the content of oral histories truly accessible to the public. Oral history interviews are often quite long. Given only a summary description of an interview's content, a listener must either dedicate an hour or two to listening to an interview in its entirety or use audio player controls to tediously jump around in a file looking for parts of interest. Furthermore, interviews varying in length typically are given descriptions of equivalent extent, meaning that long interviews are less ``densely'' described. Interviews can be made more accessible to the public by dividing them into shorter segments that can be consumed more easily and described more comprehensively. Lambert and Frisch \cite{Lambert2012} advocate for segments of five to fifteen minutes in length, with boundaries at ``a natural ending of a topic, a break before a new question, or other natural pause in the flow of speech.'' Making such segments, rather than whole interviews, the target of archival description ``unifies the scale of the navigable unit'' \cite{Lambert2012} and solves the problem of longer interviews being less comprehensively described than shorter ones.

Oral histories have a recognizable structure: a sequence of speaking turns taken by the interviewer and respondent. The possibility of identifying topical segments within this structure is implicit in the extractions oral historians make from their interviews when presenting their work to others. In some cases, these topical segments may directly correspond to speaking turns, as when an interviewer asks a topical question and the respondent answers it completely and without digression in a single turn. But usually speaking turns do not map neatly to topics: topics either stretch across several turns, or a single long response encompasses more than one topic. Interviewers may introduce topics, but respondents have their own agendas and may or may not follow where the interviewer leads. Despite these complications, historians are able to make topical segmentation decisions. If these decisions are grounded in evidence given in the interviews themselves (e.g. the language used), then one might expect there to be a certain degree of consensus among the decisions made by different individuals.

\section{Discourse Segmentation}

In the previous section I characterized a topical segment in a pragmatic way as a salient, coherent extract from an oral history. But the more general notion of a topical segment is harder to characterize, as \textit{topic} is not a clearly defined concept \cite[17]{Stede2011}. Intuitively, a topic is something that some passage of text is ``about,'' but given a passage that is judged to be about a topic $A$, one can often discern within that passage subtopics $A_{1}$, $A_{2}$, and so on---right on down to the level of individual utterances. Topical structure is hierarchical \cite{Manning1998}. Ashplant \cite[107]{Ashplant1998} demonstrates that this is true of oral histories in his analysis of part of \textit{The Dillen} \cite{Hewins1981}. He discerns a three-level topical hierarchy, with \textit{anecdotes} grouped into \textit{narrative elements}, which are in turn grouped into broad topical \textit{clusters}. Unfortunately, hierarchies can be computationally intractable, so it is preferable to focus on a specific level in the topical hierarchy that is appropriate for the task at hand, making the topical segmentation a flat, non-overlapping partitioning of the text \cite[17]{Stede2011}. For the specific task of selecting extracts from oral histories, I focus on segments in the middle of the topical hierarchy, equivalent to Ashplant's \textit{narrative elements}.

The vast majority of work on processing transcripts to segment recordings of speech has focused on broadcast news (e.g. \cite{Beeferman1999}) or conversational dialogue in settings such as phone conversations and meetings (e.g. \cite{Galley2003}). But few studies have focused specifically on segmentation for the purpose of identifying potential extracts from oral histories. Franz et al. \cite{Franz2003}, working with a collection of oral history interviews with Holocaust survivors, trained a classifier for recognizing segment boundaries, using as training data the segment boundaries defined by catalogers. Using a training set of approximately 1.5 million words (about 2800 segments), they achieved an equal error rate of 23\% on a manually transcribed test set. In a follow-up study using a subset of the same corpus, Zhang and Soergel \cite{Zhang2006} did a qualitative coding of questions and cataloging segments, and explored using these codes as features in a classifier for recognizing topical segment boundaries, with inconclusive results.

A drawback of treating topic segmentation as a supervised classification problem is that it requires as training data a corpus of texts that have been manually annotated with topic boundaries. Generating such data is expensive. This expense is increased by the fact that judgments about topic change are largely made intuitively, resulting in low inter-annotator agreement, so finding a consensus on boundary judgments may require several annotators for the same text \cite[18]{Stede2011}. Thus most research has focused on finding unsupervised techniques for segmenting topics.

The assumption behind unsupervised topic segmentation of a continuous text is that there will be greater similarity in the use of language \textit{within} topically coherent segments than \textit{across} such segments. So the first question to be answered is how ``similarity in the use of language'' will be defined and quantitatively measured. Once that is answered, a continuous text can be broken into pieces (such as individual sentences) and the similarity of those pieces to one another can be measured. This will produce an overall map of the similarity of sub-segment units in the text. The next step is to use this map to group these units into larger topical segments. This requires answering a second question: given some measure of similarity among the analyzed units, how should one decide whether a topical boundary separates them? Thus any technique for topically segmenting continuous text can be characterized in terms of how it addresses two sub-problems: measuring language similarity and postulating boundaries \cite[30]{Stede2011}.

A straightforward way of measuring language similarity is to directly compare the words used. The TextTiling algorithm \cite{Hearst1997} treats regular chunks of text as term vectors and measures the cosine similarity between them. Hearst demonstrated that it worked well for segmenting the expository prose found in a popular science article. An alternative to directly comparing word distributions is to compare higher-level semantic or statistical representations of the text, so that different but related words may contribute to the similarity measure. One such representation is a probabilistic language model, which assumes that texts are generated according to (unknown) probability distributions over topics and words. The BayesSeg algorithm \cite{Eisenstein2008} uses such a generative language model to formalize the notion of language similarity: similar texts are ones produced by the same lexical distribution. Intuitively, one might expect oral history respondents to use a less consistent vocabulary than professional writers of expository texts such as news articles, which suggests that a similarity measure based on a higher-level representation of language may be more effective for segmenting oral history transcripts.

Once there is a way of measuring language similarity between different parts of a text, one can turn to the problem of postulating topic segment boundaries. One strategy is to compare the similarity of adjacent chunks of text, looking for ``valleys'' (local minima) in a plot of the similarity scores. This is the approach taken by the TextTiling algorithm. Again, one might expect this approach to work well for professionally-produced texts with clear shifts in topic. But Malioutov and Barzilay \cite{Malioutov2006} demonstrated that only examining adjacent chunks performs less well when topic shifts are more subtle. Working with transcripts of classroom lectures, they obtained better segmentation performance by comparing (nearly) all pairs of chunks in a text, enabling them to find longer-range similarities in language use. BayesSeg \cite{Eisenstein2008} also finds a globally optimal solution, taking advantage of its probabilistic model to find the most likely segmentation. Because topic shifts in oral histories are gradual and nonlinear, a boundary computation approach that only compares adjacent blocks should be outperformed by one that can take the entire text into account.

\section{Data and Methods}

Our corpus consisted of 19 transcripts of oral history interviews conducted by the Southern Oral History Program (SOHP) at the University of North Carolina.\footnote{All of the data and code discussed in this paper are available at \url{https://github.com/contours}.} In an earlier project, SOHP staff transcribed and selected salient extracts from each of the interviews, producing TEI XML files containing the transcribed text with milestones indicating the start and end of each extract. The milestones thus divide each interview into segments, of which some subset (the selected extracts) have been judged to be topically coherent. On average, half of the segments were selected as salient extracts. The unselected segments also tended to be longer, with a mean length of 70 sentences compared to 44 sentences for the selected extracts. This suggests that the extraction process may not identify some potential topic boundaries (those that appear within the segments of interviews judged to be less salient).

I then asked two non-expert annotators to imagine that they had been tasked with curating an online collection of oral histories and to select ``the most important parts'' from each transcript. Each transcript was presented as an HTML page showing only the names of the speakers and the transcribed text of their speech. The annotators could click on the text to split it into segments and then indicate which segments were to be selected as extracts. They were instructed that each extract ``should cover a single topic or anecdote and should be understandable on its own.'' To give them a sense of the expected granularity of the extracts, the annotators were told that the length of extracts could vary considerably but would average around 30--50 sentences (the average extract length in the original project).\footnote{The complete text of the instructions provided to the annotators is available at \url{https://github.com/contours/segment/blob/5404fce/public/instructions.html}.} Extracts could not overlap, and not all of the text had to be extracted (i.e. it was permissible to ``leave out'' parts of the transcript between extracts). Extract boundaries were not limited to speaker changes or paragraph breaks and could be placed between any two adjacent sentences.

Rather than evaluating the segmentation algorithms with the best published results, I sought to evaluate algorithms representing different solutions to the two problems of measuring language similarity and postulating topic boundaries. TextTiling \cite{Hearst1997}, as described above, uses the cosine metric to quantify text similarity, and a local valley-finding algorithm to postulate boundaries. BayesSeg \cite{Eisenstein2008} probabilistically models similar text similarity as the product of a sparse lexical distribution, and uses a global expectation-maximization approach to postulate boundaries. I hypothesized that BayesSeg would better handle the inconsistent vocabulary and gradual topic shifts found in oral histories. To test this hypothesis, new implementations of both these algorithms were developed and the segmentations they produced compared to the manually-created segmentations.

Judgments about topic change are largely intuitive, which makes evaluation of topic segmentation techniques tricky. There are two main problems. The first problem is that it is difficult to produce a ``gold standard'' to evaluate against, or even to define what such a standard might be, in the absence of any clear rules to establish when a topic change has occurred. A common strategy for producing a gold standard is to have several human annotators segment the same texts and then try to determine the ``consensus'' segmentation by majority opinion. A second problem is that even given an accepted segmentation for comparison, there remains the issue of how to compare candidate segmentations to the accepted one. Most researchers have taken the approach of moving a window through both a gold standard segmentation and a segmentation to be evaluated, and then counting how many times the boundary counts postulated by the two algorithms within the window differ. This approach has various problems, including variance in how different kinds of errors are penalized and sensitivity to an arbitrary window size parameter \cite{Fournier2013b}. To avoid these issues I used the \textit{boundary edit distance} metric proposed by Fournier \cite{Fournier2013a}, which does not require any parameters and does not rely on a gold standard segmentation.
￼
\section{Comparing Manual Segmentations}
\label{comparing-manual-segmentations}

\begin{figure}
  \begin{center}
    \includesvg{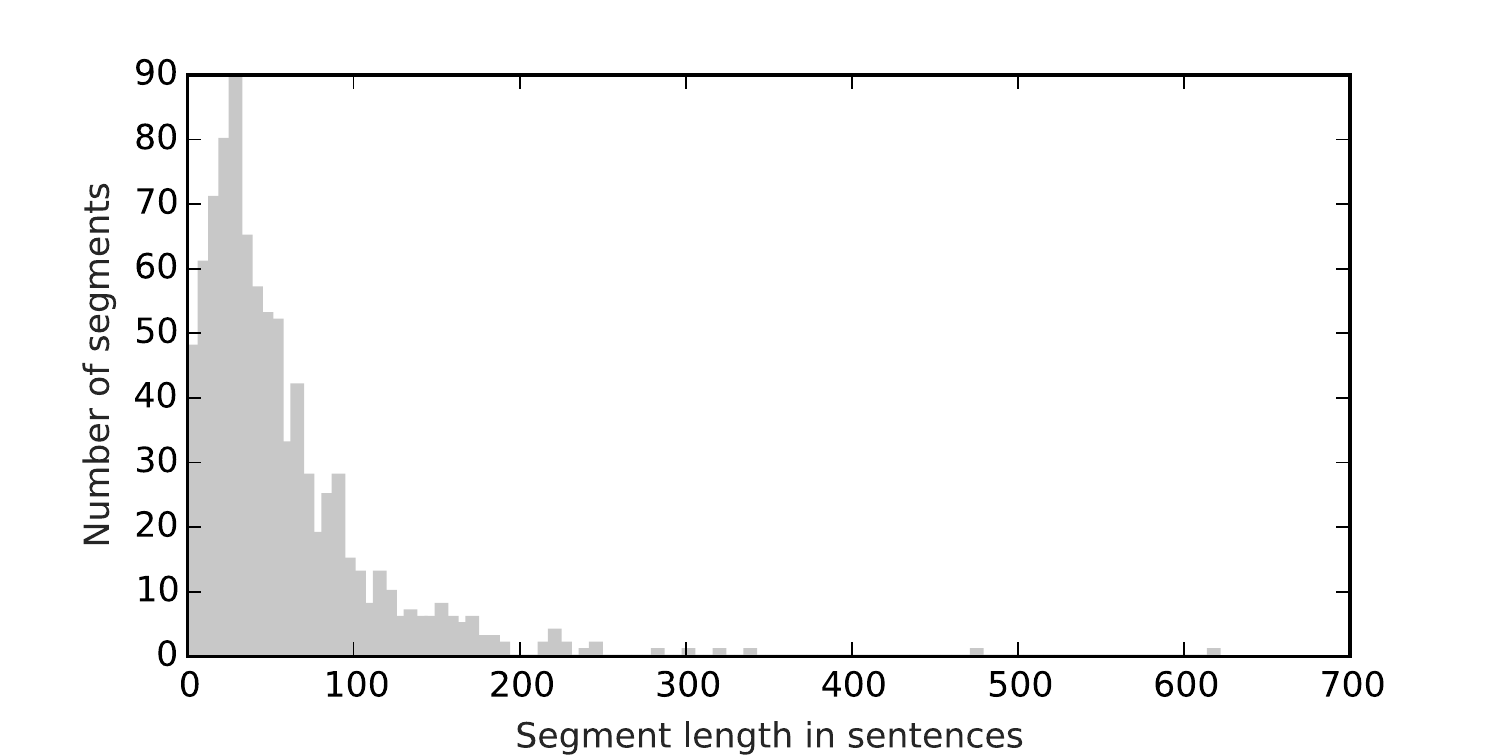}
  \end{center}
  \caption{Lengths in sentences of the manually-created segments.}
	\label{segment-length-in-sentences}
\end{figure}

The three annotators placed a total of 829 boundaries between sentences of the 19 interview transcripts, creating a total of 886 segments. Figure~\ref{segment-length-in-sentences} shows the distribution of segment lengths, which, like the topic segmentation datasets examined by Fournier \cite{Fournier2013b}, appears to be exponential. The mean segment length was $50 \pm52.7$ sentences ($n=886$). The shortest and longest segments created were one sentence and 621 sentences long, respectively. The longest segment may be an error, since it is far longer than any other segment in the dataset and far longer than the other segments created by that annotator for that interview. Some of the very short segments seem to be cases of ``trimming'' behavior, where a short segment is created specifically to be excluded (e.g. an interviewer unsuccessfully trying to interrupt an interviewee with a new question) or highlighted (e.g. a particularly vivid single quote that makes sense in isolation). Different tendencies to ``trim'' like this may explain some of the variations observed across the annotators' segmentations.

If the annotators had consistently segmented at the same level of granularity, one would expect to see little variance in the segment lengths and a positive correlation between each transcript's length and the number of segments into which it was divided. As figure~\ref{segment-length-in-sentences} shows, segment lengths varied significantly. While there was a positive correlation between transcript length and the number of segments, there was also a positive correlation between transcript length and the median segment length. Thus while the annotators divided longer transcripts into more segments, they also created longer segments for longer transcripts. Longer interview transcripts might reflect the fact that some interviewees are more long-winded than others. If that were the case then one might expect longer segments for longer interviews: each topic takes longer to cover. However it is also possible that longer interview transcripts simply cover more topics than others. If so, then the division of longer interviews into longer segments may have been due to annotator exhaustion, resulting in interviews segmented at different levels of granularity.

Segmentation granularity also varied across annotators. The original extractor placed boundaries slightly more frequently than the overall rate, which was about one boundary per 59 potential boundaries. Annotator A placed fewer boundaries (creating longer segments) on average than either the original extractor or annotator B. Annotator B placed more boundaries (creating shorter segments) on average than the other two. This could indicate that annotator B engaged in more ``trimming'' than the other annotators. These differences in boundary placement rates across annotators might lead one to expect low inter-annotator agreement, and indeed that is the case.

To measure inter-annotator agreement I used the \textit{boundary edit distance} metric proposed by Fournier \cite{Fournier2013a}. This metric compares two segmentations by examining pairs of boundary placed and treating each pair as a ``match'' (both segmentations placed a boundary at the same point), a ``near miss'' (the two segmentations placed boundaries close to one another but not at the same point), or a ``miss'' (only one of the two segmentations placed a boundary). Matches are given a score of one, misses a score of zero, and scores for near misses are scaled by how close the nearly missing boundaries are. The combined scores are normalized so that segmentation comparisons consisting of all misses will have a boundary similarity score of zero, and segmentations consisting of all matches will have a score of one. How near a pair of boundaries must be to be considered a ``near miss'' rather than a miss is determined by a transposition spanning distance parameter $n_{t}$: if the boundaries are within $n_{t}$ potential boundaries of one another, they are a near miss.

Fournier \cite[152]{Fournier2013b} suggests that for comparing segmentations of texts at paragraph boundaries, setting $n_{t}$ to 2 is appropriate, so that only pairs of boundary judgments that differ by one paragraph (and no more) will be considered ``near misses.'' Interview transcripts, being transcriptions of spoken language, are divided not into paragraphs but into speaker turns. However our units of segmentation are sentences, not speaker turns, as a topic change might be judged to occur mid-turn (i.e. when a speaker changes the subject). The mean interviewee turn length in sentences was 8, so I set $n_{t}=9$ so that pairs of boundary judgments differing by eight sentences or less would be treated as ``near misses.'' With $n_{t}=9$, the micro-averaged edit distance between pairs of boundaries placed by two annotators was $0.27 \pm0.0232$ (95\% CI, $n=1270$). This pairwise mean boundary edit distance measures actual agreement; to correct for chance agreement one can calculate Fleiss' $\pi^{\ast}$ coefficient \cite{Fleiss1979}, which was also 0.27. The closeness of agreement corrected for chance and actual agreement indicates that expected agreement due to chance is quite low. This is as expected given that were few boundaries placed compared to the number of potential boundaries.

\section{Comparing Automatic Segmentations}

All 19 transcripts were automatically segmented using our implementations of the TextTiling and BayesSeg algorithms. Stopwords were removed (using Choi's stopword list\cite{Choi2000}) and tokens were stemmed before segmentation. Before running the BayesSeg algorithm, the tokens were further filtered by POS-tagging them using the Stanford POS tagger \cite{Toutanova2003} and removing all but the nouns. The TextTiling algorithm has a number of parameters to be set. I used the same defaults as \cite{Hearst1997} (token-sequence size of 20, block size of 10, one round of smoothing, and smoothing width of 2), and the ``liberal'' boundary threshold of one standard deviation from the mean depth score. BayesSeg requires that the number of desired segments be specified, so I set the number of desired segments for each transcript to be equal to the number of segments created for that transcript by the original extractor.

I evaluated the two automatic segmentation algorithms using the approach advocated by Fournier \cite{Fournier2013b}. The segmentations produced by each algorithm were compared to all of the manual segmentations. Because boundary edit distance can be calculated at the level of individual pairs of boundaries, one can evaluate a segmentation algorithm by comparing each boundary it placed to the closest boundary placed by each of the human annotators, and then taking a micro-average of the comparison scores. To put these micro-averaged scores into context, I established lower bounds and upper bounds on performance by evaluating a ``uniform'' segmentation algorithm (lower bound) and treating the original segmentations as if they had been algorithmically produced (upper bound). The ``uniform'' segmentation algorithm produced segments having a uniform length equal to the median length of the manually-created segments. The number of observations is each group varied due to the fact that different algorithms placed different number of boundaries, resulting in different numbers of possible pairs for comparison.

\begin{figure}
  \begin{center}
    \includesvg{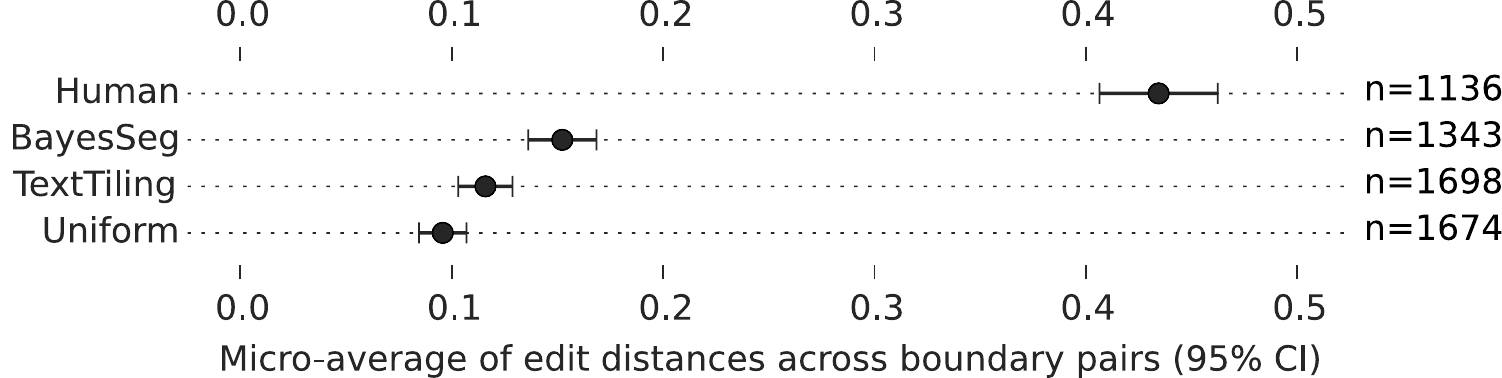}
  \end{center}
  \caption{All differences are significant except between TextTiling and Uniform.}
	\label{micro-average-of-edit-distances-across-boundary-pairs}
\end{figure}

Figure~\ref{micro-average-of-edit-distances-across-boundary-pairs} shows the micro-average score across boundary judgment pairs for each of the four groups. The number of observations is each group varied due to the fact that different algorithms placed different number of boundaries, resulting in different numbers of possible pairs for comparison. The scores are not normally distributed, so a Kruskal-Wallis distribution-free test \cite[190--199]{Hollander1999} was performed to establish that the scores differed significantly among all the groups ($H=329, 3 d.f., \alpha=0.05$). Pairwise multiple comparisons among groups using the Dwass, Steel, Critchlow-Flinger procedure \cite[240--248]{Hollander1999} established that there were significant differences between each pair except TextTiling and the uniform segmenter ($W^{*}=0.891, \alpha=0.05$). In particular, the difference between BayesSeg and TextTiling was small but significant ($W^{*}=3.83, \alpha=0.05$).

\begin{figure}
  \begin{center}
    \includesvg{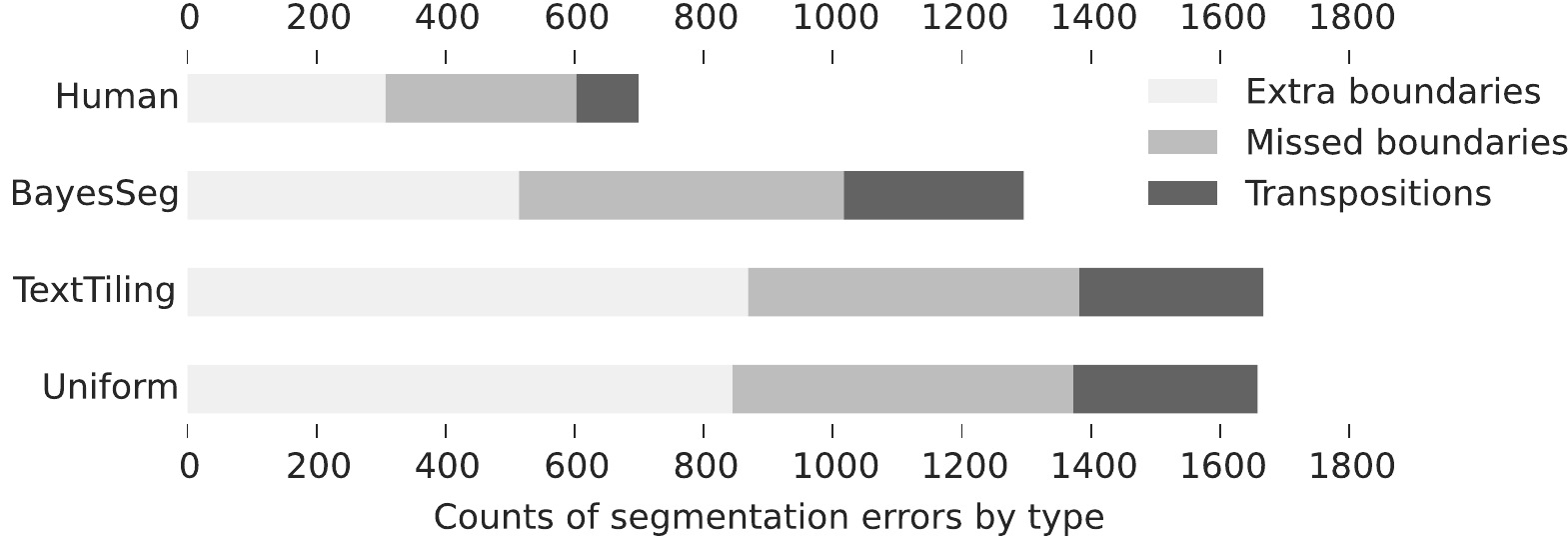}
  \end{center}
  \caption{TextTiling suffered compared to BayesSeg due to the number of extra boundaries it placed.}
	\label{counts-of-segmentation-errors-by-type}
\end{figure}

Figure~\ref{counts-of-segmentation-errors-by-type} compares the types of error made by the various segmenters. TextTiling suffered compared to BayesSeg due to the number of extra boundaries it placed (false positives). Setting a more conservative depth score threshold for the TextTiling algorithm improved precision (fewer false positives) but at the expense of recall (more false negatives), resulting in worse overall performance.

Comparing the boundary edit distances between the automatic segmenters and each individual annotator showed no significant differences between annotators ($H=4.67, \alpha=0.05$), indicating that the automatic segmenters did not agree significantly more with any one annotator. A comparison of the boundary edit distances between the automatic and manual segmentations of each interview showed significant differences only between interview U-0012 and most of the other interviews. Interview U-0012 had the second-highest level of agreement among the human annotators and the highest ratio of manually placed boundaries to potential boundaries, which may help account for the automatic segmenters' better performance.

\section{Discussion and Future Work}

Segmentation and segment-level description could make oral histories more accessible. But even though identifying topically coherent segments within interviews is an accepted part of working with oral histories, people often do not agree on the boundaries of those segments. In this study, annotators agreed (exact match or ``near miss'') on less than half of the boundaries they placed. Despite this low agreement, there is some evidence that automatic segmentation algorithms can produce segmentations that agree with human judgment better than uniform segmentations do. In particular, the BayesSeg algorithm's richer language model and globally optimal boundary selection \cite{Eisenstein2008} seem to improve its precision when segmenting oral history transcripts characterized by varied diction and gradual topic shifts. Further progress will depend both on clearer definition of the segmentation task, and on evaluation of whether automatic segmentations are adequate for building working systems for organizing oral history.

To improve agreement, the task of choosing topic boundaries in oral histories should be further constrained. To address the issue of varying segmentation granularity, annotators should be given a fixed number of boundaries to place per transcript, so that they are choosing not how many boundaries to place but only the locations of those boundaries. For example, the task of segmenting a two hour interview transcript could be reformulated as choosing how to distribute 11 segment boundaries (assuming a target granularity of ten minutes per segment). The task could be further constrained by setting upper and lower limits on segment size (for example a minimum of of five minutes and a maximum of fifteen). Finally, rather than characterizing the task as one of choosing extracts, which may lead to confounding ``trimming'' behavior, annotators should simply be asked to place the boundaries at major topic shifts.

Ultimately, the value of automatic segmentations of oral histories will be determined not by their quantitative agreement with manual segmentations but by their usefulness in actual working systems. How ``good'' do automatic segmenters need to be to help curators prepare oral histories for public exhibition? What kinds of interfaces can be built to take advantage of segmented content? How do different segmentation strategies help listeners and readers find topics of interest in a long interview? Answering these questions will require new publishing tools that incorporate segmentation into the production workflow of oral history projects, and new finding aids that make the segment, rather than the interview, the primary unit of description and discovery.

\section{Acknowledgments}

This work was funded by the Institute of Museum and Library Services. Thanks to the Southern Oral History Program for making the oral history interview transcripts available, Greg Maslov for his TextTiling implementation, Jacob Eisenstein for publishing the BayesSeg source code upon which my implementation was based, and Kathy Brennan and Sara Mannheimer for their annotation work.

\bibliographystyle{splncs03}
\bibliography{references}

\end{document}